\documentclass[conference]{IEEEtran}
\IEEEoverridecommandlockouts

\usepackage{algorithm}
\usepackage[noend]{algpseudocode}
\usepackage{epstopdf}
\usepackage{gensymb}
\usepackage{cite}
\usepackage{amsmath,amssymb,amsfonts}
\usepackage{subcaption}
\usepackage{caption}
\usepackage{graphicx}
\usepackage{textcomp}
\usepackage{xcolor}
\def\BibTeX{{\rm B\kern-.05em{\sc i\kern-.025em b}\kern-.08em
    T\kern-.1667em\lower.7ex\hbox{E}\kern-.125emX}}

\IEEEoverridecommandlockouts\IEEEpubid{\makebox[\columnwidth]{ 978-1-6654-2849-1/21/\$31.00~\copyright~2021 IEEE \hfill} \hspace{\columnsep}\makebox[\columnwidth]{ }}

\begin{document}

\title{Fast Obstacle Avoidance Motion in Small Quadcopter operation in a Cluttered Environment\\
\thanks{Chaitanyavishnu S. Gadde, Mohitvishnu S. Gadde, Nishant Mohanty, and Suresh Sundaram are with the Department
of Aerospace Engineering, Indian Institute of Science, Bangalore,  India. (e-mail: n140993@rguktn.ac.in, (mohitvishnug, nishantm, vssuresh)$@$iisc.ac.in).}
}

\author{Chaitanyavishnu S. Gadde, Mohitvishnu S. Gadde, Nishant Mohanty, and Suresh Sundaram}

\maketitle

\begin{abstract}
The autonomous operation of small quadcopters moving at high speed in an unknown cluttered environment is a challenging task. Current works in the literature formulate it as a Sense-And-Avoid (SAA) problem and address it by either developing new sensing capabilities or small form-factor processors. However, the SAA, with the high-speed operation, remains an open problem. The significant complexity arises due to the computational latency, which is critical for fast-moving quadcopters. In this paper, a novel Fast Obstacle Avoidance Motion (FOAM) algorithm is proposed to perform SAA operations. FOAM is a low-latency perception-based algorithm that uses multi-sensor fusion of a monocular camera and a 2-D LIDAR. A 2-D probabilistic occupancy map of the sensing region is generated to estimate a free space for avoiding obstacles. Also, a local planner is used to navigate the high-speed quadcopter towards a given target location while avoiding obstacles. The performance evaluation of FOAM is evaluated in simulated environments in Gazebo and AIRSIM. Real-time implementation of the same has been presented in outdoor environments using a custom-designed quadcopter operating at a speed of $4.5$ m/s. The FOAM algorithm is implemented on a low-cost computing device to demonstrate its efficacy. The results indicate that FOAM enables a small quadcopter to operate at high speed in a cluttered environment efficiently.
\end{abstract}


\section{Introduction}

{R}{ecent} advancements in Unmanned Aerial Vehicle (UAV) technologies have accelerated the deployment of UAVs for various applications like visual inspection \cite{3c854a3fe5b44f23bf0b8345d8b67db8}, reconnaissance missions \cite{5600072}, agriculture \cite{fe2adf4814da4b9b8c88ab56963bd240} etc. The current increase in humans' reliance on UAVs is due to their mechanical simplicity and ease of control. Therefore, there is a need to develop robust automation algorithms to enhance the capabilities of a UAV to operate in cluttered environments.



In the literature, the problem of obstacle avoidance has been extensively studied. Studies like \cite{1272530} have been focused on coordinated flights to prevent UAVs from crashing into each other. Similarly, \cite{Han2009ProportionalNC} uses proportional navigation for the same. Many works in UAV navigation like \cite{MANNAR2018480} use sensor data streamed to the ground station for processing, thus leading to huge latency issues. Also, in these works, a known map of the environment is used. In an uncertain environment, it will not yield successful results due to changes in the map. Therefore a sensor-based algorithm enables UAVs to act dynamically based on their observed environment.

In the literature for sensor-based obstacle avoidance, multi-sensor fusion techniques have been used to enhance the UAV's capabilities. For example, in \cite{Fasano2008MultiSensorBasedFA} one radar, four cameras, and two onboard computers have been used. However, it is not a viable option for low-cost fast moving drone operating in a cluttered environment. To overcome this problem, many algorithms using complex local \cite{Daftry2016RobustMF, 7487284, 7989677, Florence2016IntegratedPA} and global \cite{Oleynikova2016ContinuoustimeTO} planners have been designed for obstacle avoidance. However, these algorithms have not shown promising results because their planning modules were constrained due to the sensor's field of vision. Recently a state-of-the-art approach for UAV control with aggressive maneuvers has been presented by \cite{Tordesillas2019RealTimePW}, which could be used for real-time applications. However, these methods require accurate instantaneous state feedback, which can only be delivered by motion capture systems and perform well in indoor environments only. Thus, there is a need to develop a cheap multi-sensor approach to efficiently use the available sensor data for instantaneous decision-making and control in outdoor environments. 

Recently in literature, the fusion of laser scanners and RGB-D cameras has been used for obtaining accurate estimates of the environment. In \cite{Bachrach2012EstimationPA},  RGB-D sensors is used to obtain point-cloud data of the environment but these sensors are restricted to very low range in outdoor conditions. Accurate estimation of the obstacles can be obtained either by using stereo cameras \cite{8949363} or monocular cameras by creating depth maps using various Simultaneous Localization and Mapping (SLAM) algorithms \cite{Bachrach2009AutonomousFI}. However, these works are computationally intensive for on-board computers and yield in lower control frequency in position and velocity feedback. 

In the absence of data of the entire environment, learning-based motion planners have been commonly used. Approaches related to learning motion policies and depth from input data were demonstrated by \cite{Bills2011AutonomousMF}. It used orthogonal structures of indoor scenes to estimate the vanishing point to navigate the MAV in the corridors by maneuvering towards dominant vanishing points.  Other works  \cite{Schaal1999IsIL, 6287440, Argall2009ASO} have applied various learning techniques like imitation learning to navigate a UAV. In \cite{6630809}, autonomous flight is achieved at a stretch of $3$ km, at a speed of $1.5$ m/s in a cluttered forest environment using the imitation learning method. However, in most cases, due to high computational complexity, the UAV streams data to an external computer and then receives higher-level control commands.
Hence, there is still a need to develop a cost-effective, less computationally intensive obstacle avoidance algorithm for high-speed UAVs in a cluttered environment.

In this paper, we propose a novel Fast Obstacle Avoidance Motion (FOAM) algorithm for sense-and-avoid operations in a cluttered environment. A low-cost monocular camera is used to capture images, which are then processed to obtain the probabilistic occupancy map (POM). A 2D Lidar is also used to generate POM in horiziontal plane and is fused with that of camera images. The resultant POM is used to determine the free space by formulating it as an optimization problem. Here, local planner uses this directional information to generate high-level commands like the desired yaw and linear velocity. It enables the quadcopter to traverse the least occupied path to the desired goal. The performance of the proposed algorithm is evaluated in both simulation and actual environments. First, Gazebo simulations were carried out to verify the performance of the algorithm in a cluttered bugtrap environment. Later, simulations in the AIRSIM simulator were conducted in a forest environment to realize the near outdoor-like conditions. Finally, the proposed algorithm was tested on a custom-built quadcopter that uses Pixhawk autopilot in an outdoor environment. The results clearly indicate that FOAM is computationally less intensive and can handle obstacle at a speed of $4.5$m/s efficiently.

The remaining paper’s organisation is as follows: Section II presents the  the formulation of the problem, followed by the description of the proposed algorithm. Section III describes the simulations carried out along with the experimental setup followed by the conclusion.

\section{Fast Obstacle Avoidance Motion (FOAM) Algorithm}

In this section, the problem definition of sense-and-avoid using a high speed quadcopter is provided. The following subsection describes the generation of probability occupancy map (POM) based on monocular camera, followed by POM based on 2D lidar. Later, the fusion of the occupancy map is described along with the information about flight control. Finally, the FOAM algorithm is described followed by that a formal description of the same is provided. 

\subsection{Problem Definition}

\begin{figure}
	\centering
    \includegraphics[width=1\linewidth]{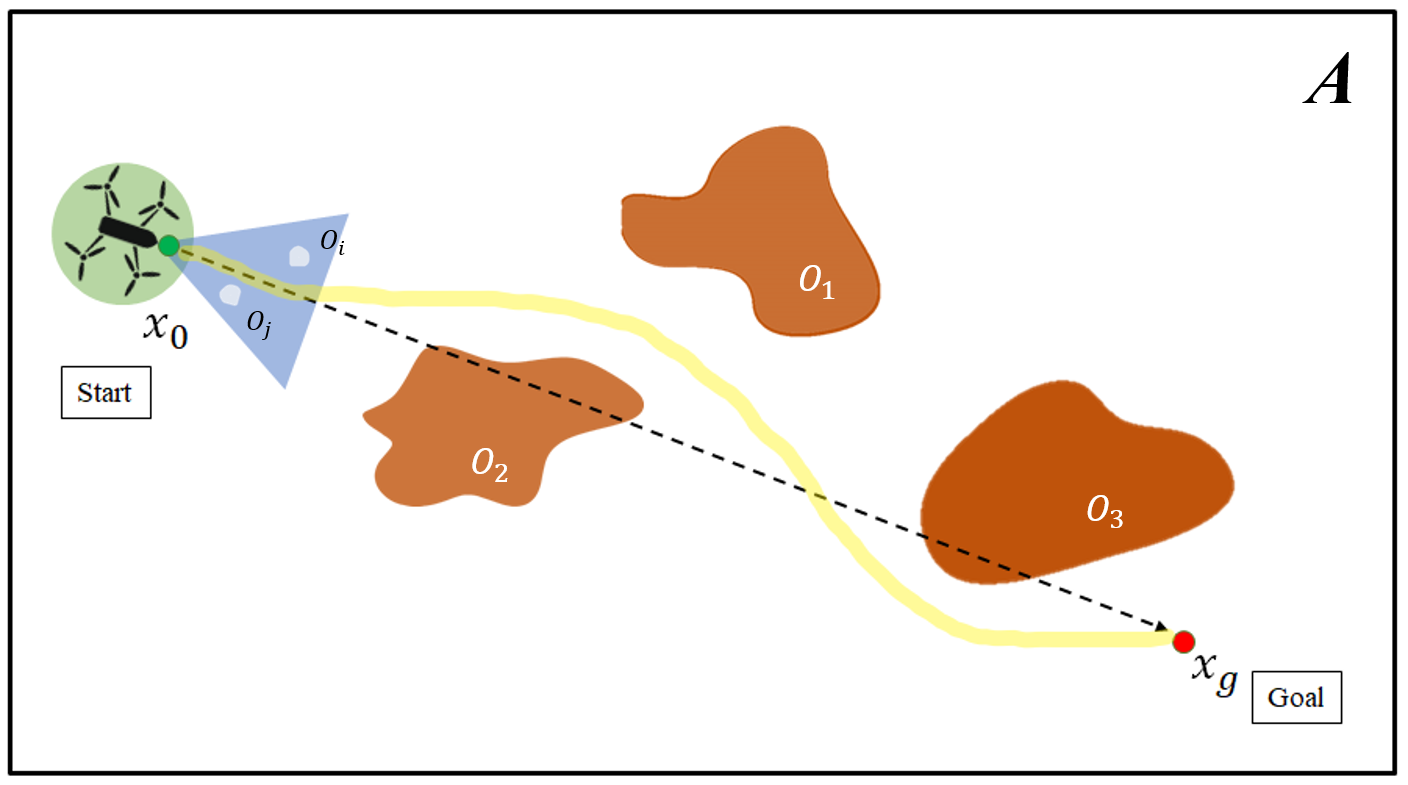}
    \caption{A typical unknown clutter environment highlighting the obstacles, launch position, goal position and sensor information. The green circle around the UAV defines the space UAV occupies in the given map. The blue region is the effective sensing zone of the UAV. The regions $O_1, O_2,…… O_i, …. O_j$ are the obstacles. The dotted line denotes the shortest path from $X_0$ (start point) to $X_g$ (goal point). The yellow path depicts a typical the UAV would follow to avoid the obstacles to reach the goal.}
    \label{fig:formuation}
\end{figure}

In this section the sense and avoid (SAA) problem is formulated for a quadcopter $Q$ with $m$ obstacles ($O_1, O_2, \cdots, O_m$) within a region $\bf{A} \subset \Re^2$. A typical SAA problem with a quadcopter and obstacles is shown in Fig. \ref{fig:formuation}. The initial launch position $({X}_o \in \Re^3)$ and the goal position $({X}_g,  \in \Re^3)$ are user defined and can be anywhere within $\bf{A}$. The objective of $Q$ is to move from the initial launch position ${x}_o$ to the goal position ${x}_g$ at a high speed $v$ (i.e, $\geq 4$ m/s). Also due to the presence of obstacles in the path it has to avoid them to achieve a collision free maneuver. The mission is assumed to be a success if $Q$ manages to reach the goal position without any collision. However the mission is terminated if the quadcopter fails to reach the goal position within $t^{max}$, or in case there is a collision. Before the start of the mission, the quadcopter is initialized at $x_o$ with velocity $v = 0$ along with a random heading angle $\psi$.Also the quadcopter is made to maintain a fixed height of $h$ along the z axis all the time. During the mission the kinematics equations for the control of the quadcopter can be given as: $\dot{x} = v*cos(\psi); \dot{y} = v*sin(\psi)$;
where, $\dot{x}$ and $\dot{y}$ represents the velocity of the $Q$ along Cartesian $x$ and $y$ axes. 

The quadcopter is equipped with a monocular camera and a 2-D lidar. The 2D lidar provides point-cloud data at a radius of $r$. Every point in the point cloud is associated with two values, i.e., the range measurements ($z_i \leq r$) and the angle from the evader frame of reference. The monocular camera provides visual sensor information within a $90 \degree$ field of view. The orientation of the camera is kept similar to that of the quadcopter. In addition to LIDAR and monocular vision, the quadcopter also has a flight controller that provides inertial measurements and position information of the vehicle. Based on the sensing information, the quadcopter is expected to identify the boundary of obstacles in the sensing region and follow an obstacle free paths.
For example in Fig. \ref{fig:formuation}, the vehicle senses the obstacle ($o_1$) partially and generate a feasible path to reach the goal point as shown in yellow coloured path. 

In order to ensure that the quadcopter reaches the goal position optimally, some  additional constraint on its motion has been added. To ensure optimally the quadcopter requires to minimize the deviation from the ideal straight line path joining the initial position and the goal while performing the mission. This has to be implemented using a local planner by minimizing the distance between the quadcopter's position $(x,y)$ from the line joining $(x_o,y_o)$ and $(x_g, y_g)$. Hence, the final mission of the quadcopter is to reach the target location by avoiding collision while having minimum deviation from the optimal path. 


\begin{figure}[h]
	\centering
    \subcaptionbox{Camera FOV divided into $s_M$ sectors\label{fig:camera_POM}}
    {\includegraphics[width=0.31\textwidth]{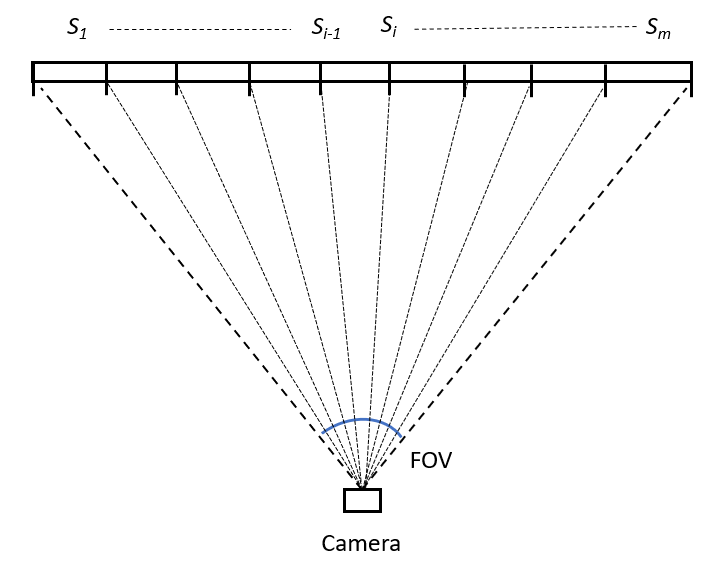}}
    \subcaptionbox{Filtered 2D lidar FOV\label{fig:lidar_pom}}
    {\includegraphics[width=0.31\textwidth]{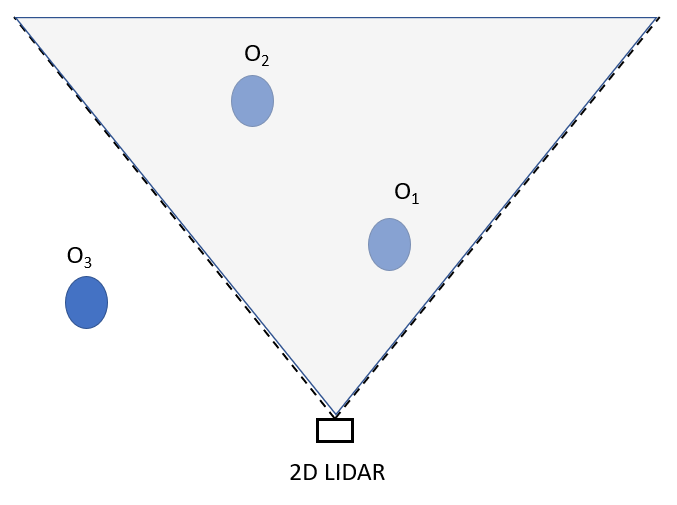}}
    \caption{Occupancy sector-based field-of-view of sensors}
    \label{fig:pom_gen}
\end{figure}


\subsection{Probabilistic Occupancy Map for FOAM}\label{pom}

The problem of detecting obstacles from the current frame $f$ at time $t$ is converted to a free space determination problem. In this case, we use a monocular camera and a 2D lidar as shown in Fig. \ref{fig:pom_gen}. Based on the field of view (FOV) of the monocular camera, the width of the image is divided into $M$ sectors, ${s_1, s_2, ..., s_M}$ (where $M$ is always an odd number) as shown in Fig. \ref{fig:camera_POM}. Similarly, the point cloud data obtained from the 2D lidar is used. Even though a 2D lidar can provide a $360^{\circ}$ field of view, the algorithm selects the region of interest which camera FOV. Hence, the point cloud data is also mapped into $M$ sectors as that of the camera, shown in Fig. \ref{fig:lidar_pom}. The objective to do so is to find the sector free from obstacles which has the minimum deviation from the median sector.  




In order to determine the free space available for the quadcopter to navigate towards the goal, we determine the probability of a sector occupied with an obstacle. A monocular camera is used to capture RGB images (30 fps). Each RGB image (640 X 480 resolution) is converted to a gray-scale image. From the gray-scaled image, robust corners are extracted using the Shi-Tomasi corner detector algorithm \cite{323794}. It is shown in the literature that these features are robust to track in an image where the changes are not abrupt. One should note that in a nominal wind condition, the objects present in the cluttered environment do not move abruptly. Hence, the corner features ($x_i$, $i=1,2,\cdots,N$) detected will be useful to identify the free space present in the current frame. Further, the optical flow values of these sparse features set are computed using the Lucas-Kanade method of iterative pyramids \cite{inproceedings}. For this purpose, we use the past three frames to track these corner features. Let $n_i$ be the number of features present in $i$th sector, $v_i$ is the flow for given feature $x_i$ and $N=\sum^m_{i=1}n_i$ be the total number of features with reliable optical flow values in the image frame. Using the flow values of these features, we compute the Probability of Occupancy Map (POM) for $i$th sector as given below,

\begin{equation}\label{Camera_POM}
P_i^c\left(I_t\right)=\frac{\sum_{j=1\ \& \ x_i\ \in s_i}^{n_i}v_j}{\epsilon+\sum_{j=1}^{N}v_j},\ i=1,2,\cdots,m
\end{equation}
where, $s_i$ is the $i^{th}$ sector and the value of $\epsilon$ is close to zero. Note that the flow values for the features detected very far away from the camera will be low, and flow values for features near the camera will be high. Hence, the occupancy value of the sector will be small if the obstacle is far away from the quadcopter and vice-versa. The computational requirement of corner features and optical flow is small, and one achieves 30 frames per second processing speed to detect the free space.

Similarly, a long range 2D lidar is used for depth estimation of the obstacles planar to the quadcopter. 
In order to estimate the occupancy in the camera plane, the field of view (FOV) of lidar is restricted to the camera's  FOV, as shown in Fig. \ref{fig:lidar_pom}. The effective depth of the 2D lidar is restricted to $d_{max}$. Based on the resolution of LIDAR, one will always have a finite number of points in each sector.  Note that the depth information of points beyond the desired sensing radius is considered effective depth. It is set at 10m in the simulation. Thus, the probability occupancy map using 2D point cloud data for the $i$th sector is given as

\begin{equation}\label{lidar_POM}
P_i^L\left(I_t\right)=\frac{\sum_{j=1\ \&\ p_i\ \in s_i}^{n_i}{d_{max}-d}_j}{\epsilon+\sum_{j=1}^{N}{d_{max}-d_j}},\ i=1,2,\cdots,m
\end{equation}

From the equation, we can conclude that the probability will be zero if the sector does not have any obstacles. 
The probability occupancy map obtained from camera and LIDAR point cloud data are fused as:

\begin{equation}\label{optimisation}
P_i\left(I_t\right)=w_CP_i^C\left(I_t\right)+w_LP_i^L\left(I_t\right),\ i=1,2,\cdots,m
\end{equation}

where $w_C$ and $w_L$ are the weights for camera and LIDAR POM respectively and $w_C + w_L = 1$.

\begin{algorithm}

\caption{Pseudocode FOAM}
\begin{algorithmic}

\Statex
 \State Data: Current image frame $f_k’$, Lidar Data $L$, GPS Data $G$;
 \State $\epsilon$ $\leftarrow$ positive value close to zero;
 \State $H$ $\leftarrow$ empty list;
 \State $P$ $\leftarrow$ empty list;
 \State Algorithm Parameters: $W_C, W_L | W_C + W_L = 2, D_{max}, M$
 
 \Function{FOAM}{} 
    \For{i in $1, 2, ……. M$}
        \State $P_i^C$ $\leftarrow$ POM $image ()$
        \State $P_i^L$ $\leftarrow$ POM $lidar ()$
        \State $P_i$ $\leftarrow$ $W_C P_i^C + W_L P_i^L$
        \State $P$ $\leftarrow$ append ($P_i$)
    \EndFor

    \State $s_d$ $\leftarrow$ minYawCost (P)
    \State currYaw $\leftarrow$ from GPS Data G
\If {$s_d$ $>$ ($\frac{M + 1}{2}$)}
\State yaw $\leftarrow$ currYaw + 10 * (i – ($\frac{M + 1}{2}$))

\ElsIf {$s_d$ $<$ ($\frac{M + 1}{2}$)}
\State yaw $\leftarrow$ currYaw – 10 * (($\frac{M + 1}{2}$) - i)

\Else
\State yaw $\leftarrow$ currYaw
\EndIf

\EndFunction

\Function{POM image} {}

\State Get $f_k’$ from camera
\State $f_k$ $\leftarrow$ convert $f_k’$ to gray-scale image
\State $X_i$ $\leftarrow$ corner features for sector $s_i$ in $f_k$ \cite{323794}
\State $V_i$ $\leftarrow$ optical flow values for given $X_i$ in frames $f_k– 1$,$f_k – 2$,$f_k-3$ \cite{inproceedings}
\State calculate $P_i^C$ using Equation (\ref{Camera_POM})
\State return $P_i^C$

\EndFunction

\Function {POM lidar}{}

\State Get $L$ from lidar
\State $D_j$ $\leftarrow$ lidar data in sector $S_i$ in frame $f_k, D_j \in L$
\State calculate $P_i^l$ using Equation (\ref{lidar_POM})
\State return $P_i^L$
\EndFunction

\Function {minYawCost (P)}{}

\State $P_{min}$ = min(P)
\For {l in 0, 1, ……. ($M$ - 1)}
\If {P = $P_{min}$}
\State H $\leftarrow$ append ($l$ + 1)
\Else
\State H $\leftarrow$ min (abs (h – ($\frac{M + 1}{2}$))) $\forall h \in H$
\EndIf
\EndFor
\EndFunction

\end{algorithmic}
\label{Algo}
\end{algorithm}

\subsection{Flight Control}

\begin{figure}[h]
	\centering
    \includegraphics[width=1\linewidth]{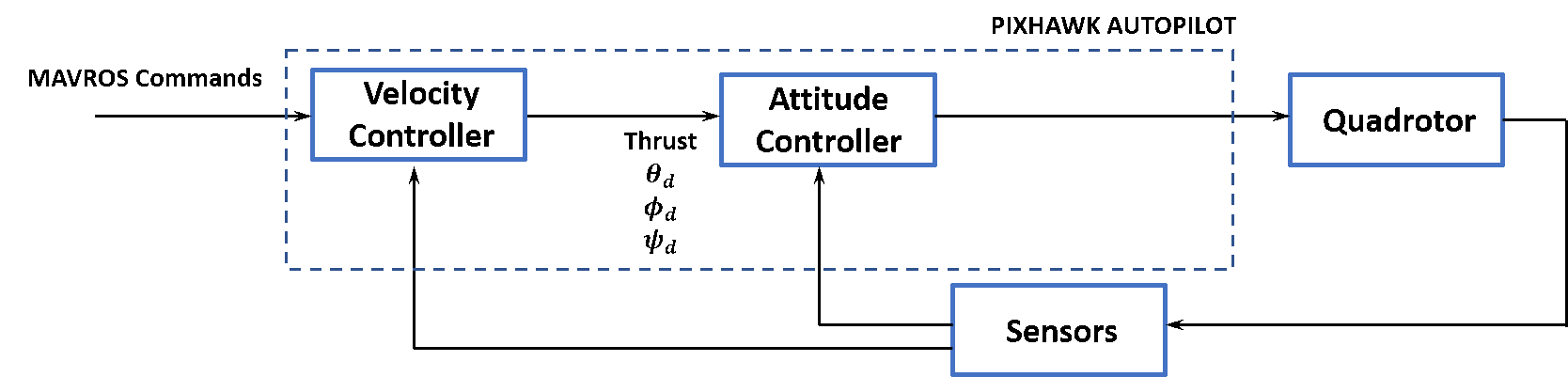}
    \caption{A Schematic Diagram of Flight Control}
    \label{fig:control}
\end{figure}

The algorithm is designed to send high-level commands to the quadrotor – (1) forward translational speed, (2) constant height, and (3) rotational velocity (heading/desired yaw). The forward velocity of the complete system is constant throughout the whole mission and can be modified by the user based on the application of the vehicle. Fig. \ref{fig:control} shows the schematic diagram of the flight control used for this algorithm. 
As mentioned, the required high-level commands like the desired yaw, height of the vehicle, etc., are sent to the velocity controller present in the Pixhawk autopilot. These velocity commands are sent via mavlink protocol while using the MAVROS package to the autopilot. The Pixhawk is connected to the on-board computer via a serial connection.
The outer loop uses a PID controller for velocity control. It computes the desired thrust, roll, pitch and yaw and sends it to the attitude controller. The PID control of the attitude controller then realises the required motor speed on the quadrotor for the optimal motion. The above-mentioned process is completely realised inside the Pixhawk autopilot of the quadcopter.

\subsection{FOAM Algorithm}

\begin{figure}
	\centering
    \includegraphics[clip,trim=0mm 0mm 0mm 0mm,width=1\linewidth]{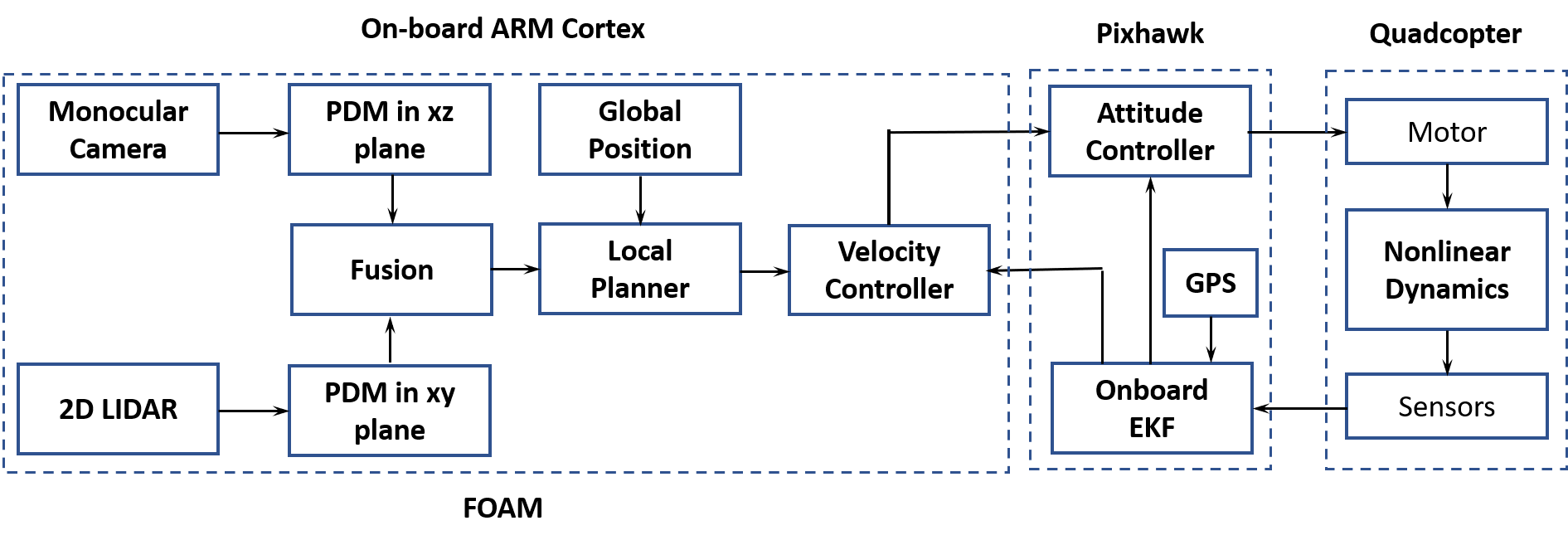}
    \caption{A Schematic Diagram of FOAM System Architecture}
    \label{fig:architecture}
\end{figure}

The FOAM algorithm is designed such that it is computationally less intensive. The system architecture of the same is shown in Fig. \ref{fig:architecture}. The figure illustrates various components of FOAM and its integration into the small quadcopter. First, FOAM generates POM based on the data from the monocular camera for the $xz$-plane. Then it generates POM using the point cloud data from 2D-LIDAR for the $xy$-plane. Next, free space is determined by formulating an optimization problem by fusing the POM’s. The solution is to determine the free space at time $t$. Finally, the local planner generates the necessary commands to the Pixhawk flight controller such that the quadcopter avoids the obstacle and move towards the goal position. The above-mentioned algorithm is summarized and given in algorithm \ref{Algo}.

\begin{figure}[h]
	\centering
    \includegraphics[width=1\linewidth]{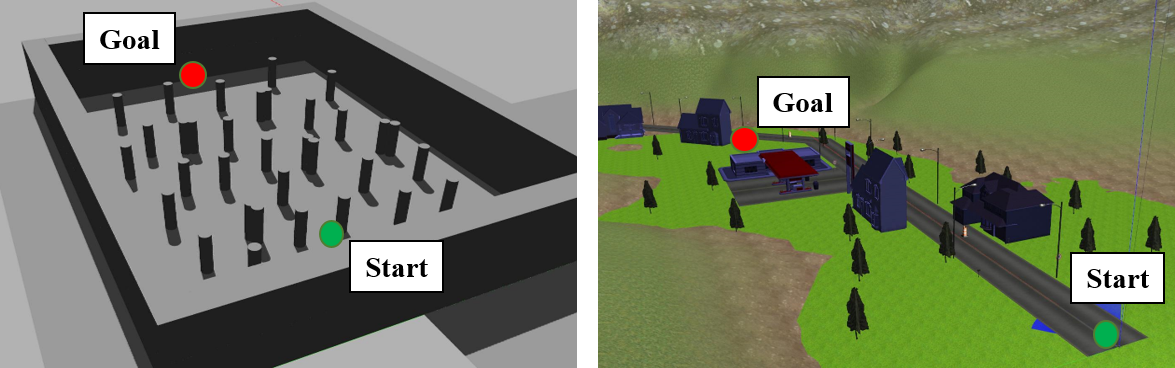}
    \caption{Bugtrap (left) and Town (right) environments used in Gazebo simulator.}
    \label{fig:gazebo}
\end{figure}

\begin{figure*}[h]
\centering
    \subcaptionbox{Simulation frame at time 26 \textit{seconds}\label{fig:a2}}
    {\includegraphics[clip,trim=0.5mm 0.5mm 0mm 0mm,width=0.33\textwidth, height=4cm]{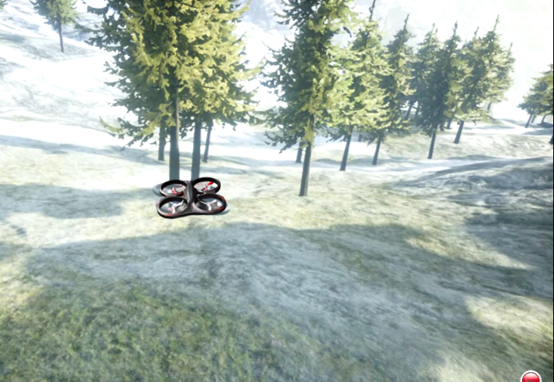}}
    \subcaptionbox{Simulation frame at time 38 \textit{seconds}\label{fig:b2}}
    {\includegraphics[clip,trim=0.5mm 0.5mm 0mm 0mm,width=0.33\textwidth , height=4 cm]{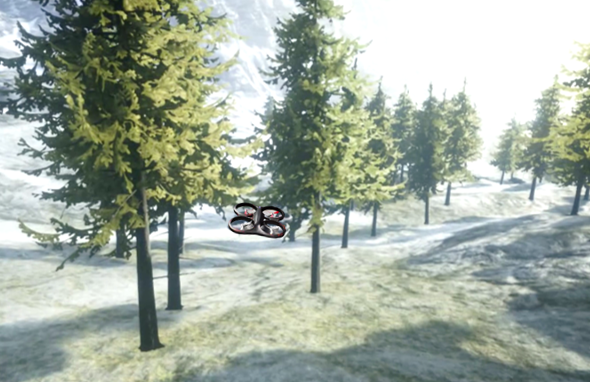}}
    \subcaptionbox{Simulation frame at time 52 \textit{seconds}\label{fig:c2}}
    {\includegraphics[clip,trim=0.5mm 0.5mm 0mm 0mm,width=0.32\textwidth , height=4 cm]{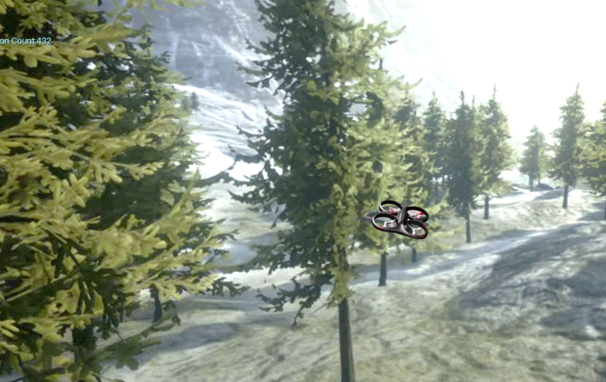}}
    
        \subcaptionbox{Experiment frame at time 66 \textit{seconds}\label{fig:d2}}
    {\includegraphics[clip,trim=1.2mm 1mm 0mm 0mm,width=0.33\textwidth, height=4cm]{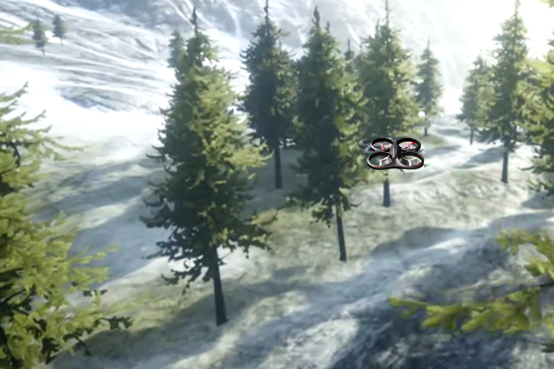}}
    \subcaptionbox{Experiment frame at time 80 \textit{seconds}\label{fig:e2}}
    {\includegraphics[clip,trim=1.2mm 1mm 0mm 0mm,width=0.33\textwidth , height=4 cm]{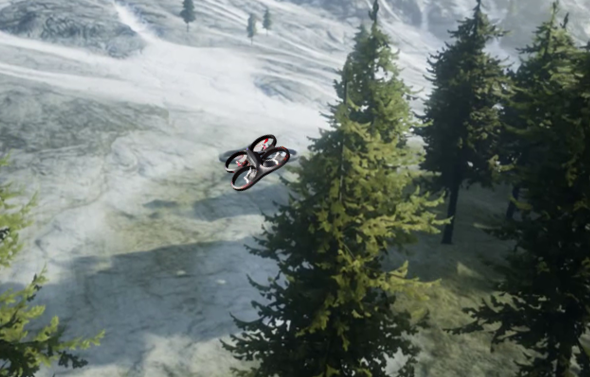}}
    \subcaptionbox{Experiment frame at time 98 \textit{seconds}\label{fig:f2}}
    {\includegraphics[clip,trim=1.2mm 1mm 0mm 0mm,width=0.32\textwidth , height=4 cm]{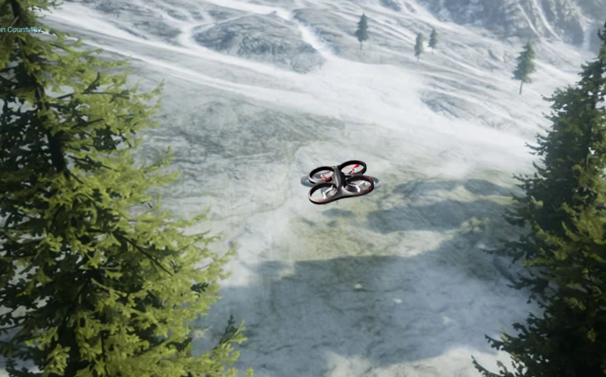}}
    \caption{Different instances of a AirSim Forest Simulation. The quadcopter flies towards Goal point while avoiding the obstacles.}
    \label{fig:sim_exp_trajectory2}
\end{figure*}

\section{Performance Evaluation}
In order to evaluate the performance of FOAM, different test scenarios are considered. Custom environments were created and tested using the Gazebo simulator using PX4 Flight Stack in the back-end for the 3DR IRIS quadcopter. Simulations were also conducted on AirSim to verify the performance of the algorithm. Finally, to evaluate the entire proposed system, experiments were conducted in an outdoor environment using a custom-built quadrotor vehicle that would autonomously navigate based on the high-level commands sent from the onboard computer. The videos related to the evaluation can be found in the following link: \textbf{https://bit.ly/3sFtVyP}.  

\subsection{PX4 ROS Gazebo Simulations}
The proposed algorithm was tested in various environments. A custom python code was designed to compute and provide the required control commands to the 3DR IRIS quadrotor. The 3DR IRIS model has a monocular camera and 2-D lidar plugins integrated into it. The monocular camera uses a horizontal field of view of $90 \degree$ and a sensing range of $10$ m for the 2-D lidar. The sample Gazebo environment is shown in Fig. \ref{fig:gazebo}.

\subsection{AIRSIM Simulation}
Next, the algorithm was tested on AirSim, an open-source simulator to realize the near outdoor-like conditions. The quadcopter model uses a built-in monocular camera along with the 2D LIDAR for perception. Different instances in Fig. \ref{fig:sim_exp_trajectory2} show a typical forest environment in an AirSim simulator. The algorithm was also tested on a custom-designed cluttered environment in AirSim to test its robustness. The simulation was carried out using Intel i7 8750H 3.0 GHz processor, powered with 16GB of DDR4 memory, which also housed an NVIDIA RTX 2070 Max-Q graphics memory.

\begin{figure}[h]
	\centering
    \includegraphics[width=1\linewidth]{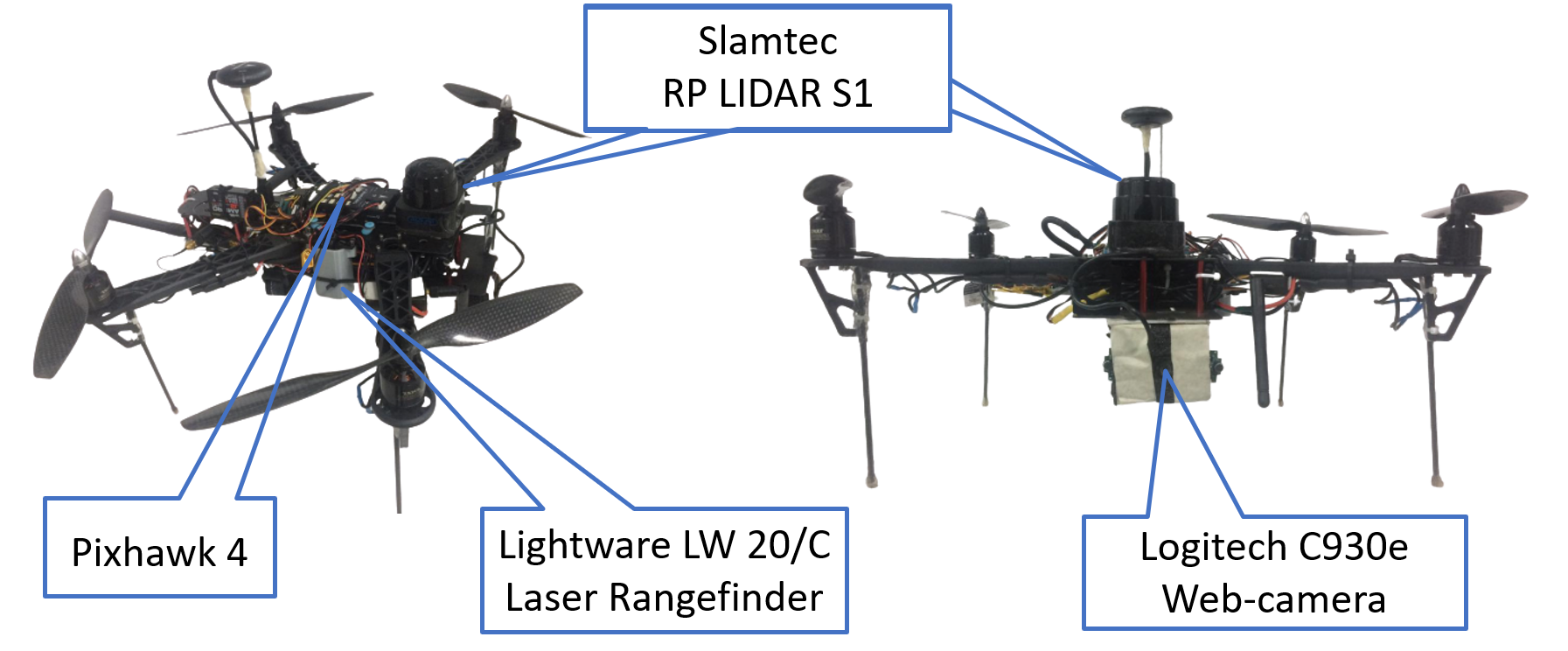}
    \caption{Custom UAV used in the experiments. It is equipped with a NVIDIA Jetson Nano, Logitech C930e web camera, RP LIDAR S1 and Lightware LW 20/c Laser Rangefinder.}
    \label{fig:quadcopter}
\end{figure}

\begin{figure}
	\centering
    \includegraphics[width=1\linewidth]{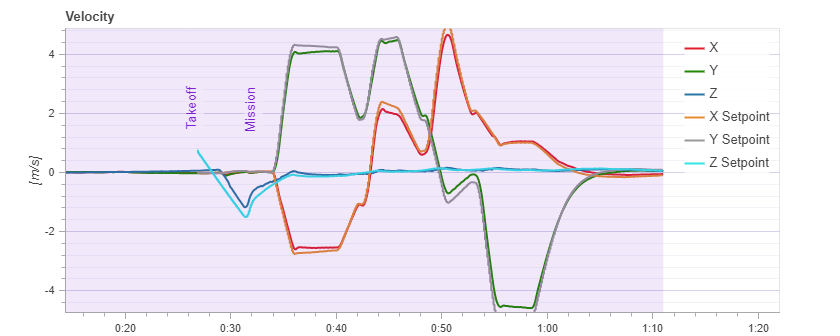}
    \caption{Plots of the velocity and tracking errors during a high speed flight at maximum speed of 4.5 m/s}
    \label{fig:control_plot}
\end{figure} 

\begin{figure*}
	\centering
    \includegraphics[width=0.9\linewidth]{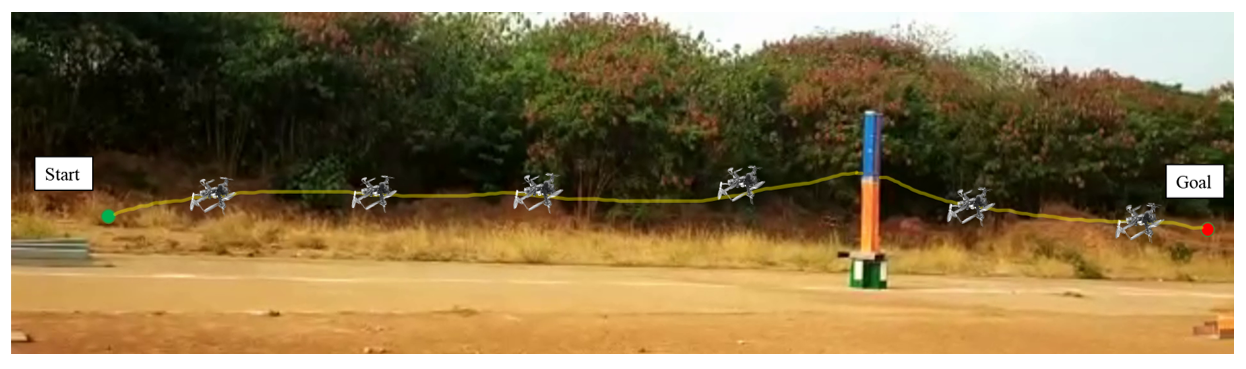}
    \caption{Composite image of the outdoor experiment. The quadcopter must travel from start    to the goal    .}
    \label{fig:outdoor}
\end{figure*}

\subsection{Outdoor Experiments}
Finally, to evaluate the proposed system for high speeds, a custom-built quadcopter shown in Fig. \ref{fig:quadcopter} was tested in an outdoor environment. The quadrotor is composed of TBS 500 Carbon fiber frame. The quadrotor houses an of the shelf ARM-based computer, NVIDIA Jetson Nano, for onboard computation of perception, control, and planning. Pixhawk 4 is the flight controller, which uses the PX4 flight stack. The vehicle has a 35 percent hover throttle on a 5000 mAh 3S Lithium Polymer (LiPo) battery. The quadcopter is equipped with a Logitech C930e web camera.  The camera module packs up with a 1080p resolution, H.264 video compression with Scalable Video Coding and UVC 1.5 encoding to minimize its dependence on computer and network resources and a wide 90-degree field of view. Slamtec RP LIDAR S1 is used for the depth estimation. It provides 360-degree scan field, 5.5hz/10hz rotating frequency with 40-meter ranger distance measurement with more than 9.2K samples per second.
A pole was erected in the middle of a field. The quadcopter was placed at the initial launch position. The destination position was given as an input and the program was run. During the mission the quadcopter was able to avoid the pole in between while operating at a desired velocity of 4.5 m/s. Also the height along z-direction was maintained, i.e, 2 m. An image instance of the same is given in fig. \ref{fig:outdoor}. Fig. \ref{fig:control_plot} shows the commanded velocity and estimated velocity. The experiment shows that FOAM can perform sense and avoid in outdoor environments at a very high speed. Also, it demonstrates the low latency of the proposed FOAM architecture by using it on low cost computing devices.

\section{Conclusion}

In this paper, the novel Fast Obstacle Avoidance Motion (FOAM) algorithm is proposed to perform sense and avoid (SAA) operations. FOAM is a low-latency perception-based algorithm that uses multi-sensor fusion of a monocular camera and a 2-D LIDAR.  It has been shown to perform obstacle avoidance at a speed of 4.5 m/s in outdoor environments. Also, it has been shown to perform high-speed maneuvers in a cluttered environment in both Gazebo and AirSim.

\bibliographystyle{IEEEtran}
\bibliography{main}

\begin{thebibliography}{10}
\providecommand{\url}[1]{#1}
\csname url@samestyle\endcsname
\providecommand{\newblock}{\relax}
\providecommand{\bibinfo}[2]{#2}
\providecommand{\BIBentrySTDinterwordspacing}{\spaceskip=0pt\relax}
\providecommand{\BIBentryALTinterwordstretchfactor}{4}
\providecommand{\BIBentryALTinterwordspacing}{\spaceskip=\fontdimen2\font plus
\BIBentryALTinterwordstretchfactor\fontdimen3\font minus
  \fontdimen4\font\relax}
\providecommand{\BIBforeignlanguage}[2]{{%
\expandafter\ifx\csname l@#1\endcsname\relax
\typeout{** WARNING: IEEEtran.bst: No hyphenation pattern has been}%
\typeout{** loaded for the language `#1'. Using the pattern for}%
\typeout{** the default language instead.}%
\else
\language=\csname l@#1\endcsname
\fi
#2}}
\providecommand{\BIBdecl}{\relax}
\BIBdecl

\bibitem{3c854a3fe5b44f23bf0b8345d8b67db8}
T.~{\"O}zaslan, G.~Loianno, J.~Keller, C.~Taylor, V.~Kumar, J.~Wozencraft, and
  T.~Hood, ``\BIBforeignlanguage{English (US)}{Autonomous navigation and
  mapping for inspection of penstocks and tunnels with mavs},''
  \emph{\BIBforeignlanguage{English (US)}{IEEE Robotics and Automation
  Letters}}, vol.~2, no.~3, pp. 1740--1747, Jul. 2017.

\bibitem{5600072}
S.~{Waharte} and N.~{Trigoni}, ``Supporting search and rescue operations with
  uavs,'' in \emph{2010 International Conference on Emerging Security
  Technologies}, 2010, pp. 142--147.

\bibitem{fe2adf4814da4b9b8c88ab56963bd240}
B.~Vroegindeweij, S.~{van Wijk}, and E.~{van Henten},
  ``\BIBforeignlanguage{English}{Autonomous unmanned aerial vehicles for
  agricultural applications},'' 2014.

\bibitem{1272530}
R.~W. {Beard} and T.~W. {McLain}, ``Multiple uav cooperative search under
  collision avoidance and limited range communication constraints,'' in
  \emph{42nd IEEE International Conference on Decision and Control (IEEE Cat.
  No.03CH37475)}, vol.~1, 2003, pp. 25--30 Vol.1.

\bibitem{Han2009ProportionalNC}
S.-C. Han, H.~Bang, and C.~Yoo, ``Proportional navigation-based collision
  avoidance for uavs,'' \emph{International Journal of Control, Automation and
  Systems}, vol.~7, pp. 553--565, 2009.

\bibitem{MANNAR2018480}
S.~Mannar, M.~Thummalapeta, S.~K. Saksena, and S.~Omkar, ``Vision-based control
  for aerial obstacle avoidance in forest environments,''
  \emph{IFAC-PapersOnLine}, vol.~51, no.~1, pp. 480--485, 2018, 5th IFAC
  Conference on Advances in Control and Optimization of Dynamical Systems ACODS
  2018.

\bibitem{Fasano2008MultiSensorBasedFA}
G.~Fasano, D.~Accardo, A.~Moccia, C.~Carbone, U.~Ciniglio, F.~Corraro, and
  S.~Luongo, ``Multi-sensor-based fully autonomous non-cooperative collision
  avoidance system for unmanned air vehicles,'' \emph{J. Aerosp. Comput. Inf.
  Commun.}, vol.~5, pp. 338--360, 2008.

\bibitem{Daftry2016RobustMF}
S.~Daftry, S.~Zeng, A.~Khan, D.~Dey, N.~Melik-Barkhudarov, J.~Bagnell, and
  M.~Hebert, ``Robust monocular flight in cluttered outdoor environments,''
  \emph{ArXiv}, vol. abs/1604.04779, 2016.

\bibitem{7487284}
{Sikang Liu}, M.~{Watterson}, S.~{Tang}, and V.~{Kumar}, ``High speed
  navigation for quadrotors with limited onboard sensing,'' in \emph{2016 IEEE
  International Conference on Robotics and Automation (ICRA)}, 2016, pp.
  1484--1491.

\bibitem{7989677}
B.~T. {Lopez} and J.~P. {How}, ``Aggressive 3-d collision avoidance for
  high-speed navigation,'' in \emph{2017 IEEE International Conference on
  Robotics and Automation (ICRA)}, 2017, pp. 5759--5765.

\bibitem{Florence2016IntegratedPA}
P.~R. Florence, J.~Carter, and R.~Tedrake, ``Integrated perception and control
  at high speed: Evaluating collision avoidance maneuvers without maps,'' in
  \emph{WAFR}, 2016.

\bibitem{Oleynikova2016ContinuoustimeTO}
H.~Oleynikova, M.~Burri, Z.~Taylor, J.~Nieto, R.~Siegwart, and E.~Galceran,
  ``Continuous-time trajectory optimization for online uav replanning,''
  \emph{2016 IEEE/RSJ International Conference on Intelligent Robots and
  Systems (IROS)}, pp. 5332--5339, 2016.

\bibitem{Tordesillas2019RealTimePW}
J.~Tordesillas, B.~Lopez, J.~Carter, J.~Ware, and J.~How, ``Real-time planning
  with multi-fidelity models for agile flights in unknown environments,''
  \emph{2019 International Conference on Robotics and Automation (ICRA)}, pp.
  725--731, 2019.

\bibitem{Bachrach2012EstimationPA}
A.~Bachrach, S.~Prentice, R.~He, P.~Henry, A.~S. Huang, M.~Krainin,
  D.~Maturana, D.~Fox, and N.~Roy, ``Estimation, planning, and mapping for
  autonomous flight using an rgb-d camera in gps-denied environments,''
  \emph{The International Journal of Robotics Research}, vol.~31, pp. 1320 --
  1343, 2012.

\bibitem{8949363}
S.~W. {Chen}, G.~V. {Nardari}, E.~S. {Lee}, C.~{Qu}, X.~{Liu}, R.~A.~F.
  {Romero}, and V.~{Kumar}, ``Sloam: Semantic lidar odometry and mapping for
  forest inventory,'' \emph{IEEE Robotics and Automation Letters}, vol.~5,
  no.~2, pp. 612--619, 2020.

\bibitem{Bachrach2009AutonomousFI}
A.~Bachrach, R.~He, and N.~Roy, ``Autonomous flight in unknown indoor
  environments,'' \emph{International Journal of Micro Air Vehicles}, vol.~1,
  pp. 217 -- 228, 2009.

\bibitem{Bills2011AutonomousMF}
C.~Bills, J.~Chen, and A.~Saxena, ``Autonomous mav flight in indoor
  environments using single image perspective cues,'' \emph{2011 IEEE
  International Conference on Robotics and Automation}, pp. 5776--5783, 2011.

\bibitem{Schaal1999IsIL}
S.~Schaal, ``Is imitation learning the route to humanoid robots?'' \emph{Trends
  in Cognitive Sciences}, vol.~3, pp. 233--242, 1999.

\bibitem{6287440}
B.~{Schölkopf}, J.~{Platt}, and T.~{Hofmann}, \emph{Boosting Structured
  Prediction for Imitation Learning}, 2007, pp. 1153--1160.

\bibitem{Argall2009ASO}
B.~Argall, S.~Chernova, M.~Veloso, and B.~Browning, ``A survey of robot
  learning from demonstration,'' \emph{Robotics Auton. Syst.}, vol.~57, pp.
  469--483, 2009.

\bibitem{6630809}
S.~{Ross}, N.~{Melik-Barkhudarov}, K.~S. {Shankar}, A.~{Wendel}, D.~{Dey},
  J.~A. {Bagnell}, and M.~{Hebert}, ``Learning monocular reactive uav control
  in cluttered natural environments,'' in \emph{2013 IEEE International
  Conference on Robotics and Automation}, 2013, pp. 1765--1772.

\bibitem{323794}
{Jianbo Shi} and {Tomasi}, ``Good features to track,'' in \emph{1994
  Proceedings of IEEE Conference on Computer Vision and Pattern Recognition},
  1994, pp. 593--600.

\bibitem{inproceedings}
B.~Lucas and T.~Kanade, ``An iterative image registration technique with an
  application to stereo vision,'' in \emph{International Joint Conference on
  Artificial Intelligence (IJCAI)}, 04 1981.

\end{thebibliography}

\end{document}